\documentclass[11pt,a4paper]{article}
\usepackage[hyperref]{acl2021}
\usepackage{times}
\usepackage{latexsym}

\usepackage{microtype}
\usepackage{amssymb,amsmath,latexsym,amsfonts,amsthm,cleveref, bm}
\usepackage{mathtools}
\usepackage{stmaryrd}
\usepackage{multirow}
\usepackage{booktabs}
\usepackage{algorithmicx,algorithm}
\usepackage{makecell}
\usepackage[switch]{lineno}
\usepackage{stfloats}
\usepackage{color}
\usepackage{graphicx}
\usepackage{subfigure}
\usepackage[noend]{algpseudocode}

\usepackage{microtype}

\aclfinalcopy % Uncomment this line for the final submission
\def\aclpaperid{1158} %  Enter the acl Paper ID here

\title{Pushing Paraphrase Away from Original Sentence: \\A Multi-Round Paraphrase Generation Approach}

\author{Zhe Lin \and Xiaojun Wan \\
 Wangxuan Institute of Computer Technology, Peking University \\
 Center for Data Science, Peking University \\
 The MOE Key Laboratory of Computational Linguistics, Peking University \\
 {\tt \{linzhe,wanxiaojun\}@pku.edu.cn} \\}

\date{}

\begin{document}
\maketitle
\begin{abstract}

In recent years, neural paraphrase generation based on Seq2Seq has achieved superior performance, however, the generated paraphrase still has the problem of lack of diversity. In this paper, we focus on improving the diversity between the generated paraphrase and the original sentence, i.e., making generated paraphrase different from the original sentence as much as possible. We propose \textbf{BTmPG} (\textbf{B}ack-\textbf{T}ranslation guided \textbf{m}ulti-round \textbf{P}araphrase \textbf{G}eneration), which leverages multi-round paraphrase generation to improve diversity and employs back-translation to preserve semantic information. We evaluate BTmPG on two benchmark datasets. Both automatic and human evaluation show BTmPG can improve the diversity of paraphrase while preserving the semantics of the original sentence. 
 
\end{abstract}
\section{Introduction}
Paraphrase generation or sentence paraphrasing is an important task in natural language processing, and it requires rewriting a sentence while preserving its semantics. Paraphrase generation has been widely used in many downstream tasks such as QA systems, semantic parsing, dialogue systems and so on. 

In recent years, deep learning techniques like sequence-to-sequence(Seq2Seq) have achieved superior performance on natural language generation tasks \citep{zhao2010leveraging,wubben2010paraphrase}.  
Many paraphrase models based on Seq2Seq have achieved inspiring results. For example, \citet{prakash2016neural} leveraged stacked residual LSTM networks to generate paraphrase, and \citet{gupta2017deep} proposed a deep generative framework based on variational auto-encoder for paraphrase generation. 

Though paraphrase generation models based on Seq2Seq have demonstrated advanced ability, the generated paraphrase still has the problem of lack of diversity, i.e., the output paraphrase only makes trivial changes to the original sentence. A good paraphrase of a sentence is one that is semantically similar to that sentence while being (very) syntactically and/or lexically different from it \citep{rahul2013squibs}. Paraphrase which is too similar to the original sentence is much less useful in many real applications. 

In this paper, we focus on improving the diversity of generated paraphrase, i.e., making generated paraphrase different from the original sentence as much as possible. An intuitive but uninvestigated idea is to adopt multi-round paraphrase generation. Concretely, we first send the original sentence into a paraphrase generation model to generate a paraphrase, and then we use the generated paraphrase as the input of the model to generate a new paraphrase. As long as we leverage a paraphrase generation model with strong diversity like variational auto-encoder (VAE)\citep{kingma2013auto}, we can get the paraphrase as different as possible from the original sentence after multi-round generation. 

However, existing paraphrase models can not ensure that the major semantics of the original sentence can be preserved after multi-round paraphrase generation, especially the model with strong diversity. With the increase of paraphrasing round, the generated sentence will be more and more different from the original sentence, and the semantics will be gradually different from the original sentence as well. To tackle this problem, we introduce back-translation to maintain the semantics of paraphrase. Back-translation, which translates the generated sentence into the original sentence, has been widely used in semi-supervised natural language generation \citep{zhao2020semi} and data augmentation\citep{li2020revisiting}, and it can improve the robustness of machine-translation system \citep{li2019improving}. We assume that paraphrase with similar semantics can be translated back to the original sentence. So, we can leverage back-translation to provide guidance for multi-round paraphrase generation.

Particularly, we propose \textbf{B}ack-\textbf{T}ranslation guided \textbf{m}ulti-round \textbf{P}araphrase \textbf{G}eneration (\textbf{BTmPG}), by combining neural paraphrase model with back-translation to generate paraphrases in a multi-round process. The contributions of our work are summarized as below:

1) We propose a new multi-round paraphrase generation method to generate diverse paraphrase that is much different from the original sentence and leverage back-translation to preserve the major semantics during the multi-round paraphrase generation.  Our code is publicly available at \url{https://github.com/L-Zhe/BTmPG}. 

2) Automatic and human evaluation results demonstrate that our method can substantially improve the diversity of generated paraphrase, while preserving the semantics during multi-round paraphrase generation.

\section{Related Work}

 Paraphrase generation or sentence paraphrasing can been seen as a monolingual translation task. 
 \citet{prakash2016neural} leveraged stacked residual LSTM networks to generate paraphrase. \citet{gupta2017deep} found deep generative model such as variational auto-encoder can be able to achieve better performance in paraphrase generation. \citet{li2019decomposable} proposed DNPG to decompose a sentence into sentence-level pattern and phrase-level pattern to make neural paraphrase generation more interpretable and controllable, and they found DNPG can be adopted into unsupervised domain adaptation method for paraphrase generation. 
 \citet{fu2019paraphrase} proposed a new paraphrase model with latent bag of words. 
\citet{wang2019task} found that adding semantics information into paraphrase model can significantly boost performance. \citet{10.1145/3394486.3403231} proposed an unsupervised paraphrase model with deep reinforcement learning framework. \citet{liu-etal-2020-unsupervised} regarded paraphrase generation as an optimization problem and proposed a sophisticated objective function.
All methods above focus on the generic quality of paraphrase and do not care about the diversity of paraphrase.

There are also some methods focusing on improving the diversity of paraphrase.   \citet{gupta2017deep} leveraged VAE to generate several different paraphrases by sampling the latent space. \citep{dips2019} provided a novel formulation of the problem in terms of monotone sub-modular function maximization to generate diverse paraphrase. \citet{goyal2020neural} used syntactic transformations to softly “reorder” the source sentence and guide paraphrase model. \citet{thompson2020paraphrase} introduced a simple paraphrase generation algorithm which discourages the production of n-grams that are present in the input to prevent trivial copies or near copies. Note that the purpose of the work \citep{gupta2017deep,DBLP:journals/corr/abs-1909-13827} is different from ours, while \citet{thompson2020paraphrase} has the same purpose with our work, i.e., pushing the generated paraphrase away from the original sentence.

\section{Model}

In this section, we introduce the components of our model in detail. First, we define the paraphrase generation task and give an overview of our model. Next, we describe the paraphrase model and the back-translation model. Then, we show how to use the gumble-softmax to connect the paraphrase model with the back-translation model. Finally, we describe the loss function and training process of our model in detail. Figure\ref{overview} shows an overview of our model. 

\subsection{Notations and Overview}

Our model regards paraphrase generation as a monolingual translation task. Given a paraphrase pair $(S_0, P)$, which $S_0$ is the original/source sentence and $P$ is the target paraphrase given in the dataset. 

\begin{figure*}[htb]
\centering
\includegraphics[scale=0.6]{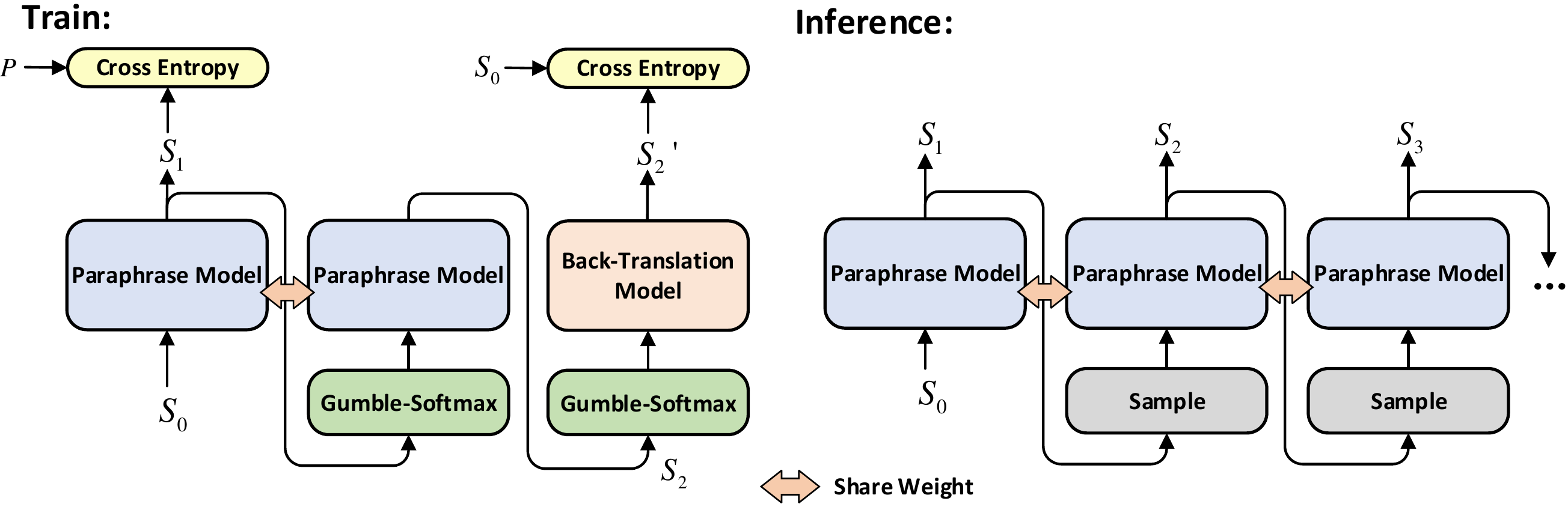}
\caption{ An overview of BTmPG, which leverages back-translation to guide paraphrase model during training and generates paraphrases in a multi-round process.}
\label{overview}
\end{figure*}

As is shown in Figure \ref{overview}, we introduce a multi-round paraphrasing method. In the \textbf{first round} generation, we send $S_0$ into a paraphrase model to generate a paraphrase $S_1$. In the \textbf{second round} generation, we use the $S_1$ as the input of the model to generate a new paraphrase $S_2$. And so forth,  in the \textbf{$i$-th round} generation, we send $S_{i-1}$ into the paraphrase model to generate $S_i$.

Although multi-round generation can increase the paraphrase diversity, the semantics of paraphrase may change during generation. We thus introduce back-translation to tackle this problem based on the assumption that paraphrase can be translated back to the original sentence while the semantic information has not been changed. In the first round, we calculate the loss between $S_1$ and $P$ to train our paraphrase model. In the $i$-th round, we send its generated paraphrase $S_i$ into a back-translation model to generate $S_i^{'}$, and we optimize the cross-entropy loss between $S_i^{'}$ and $S_0$. The back-translation model which translates the paraphrase in $i$-th round back to the original sentence can guide the paraphrase to preserve semantics during multi-round generation.

In addition, we introduce gumble-softmax embedding to tackle the problem that the model with sampling operation between different rounds' generation can not be optimized by SGD optimizer.

\subsection{Paraphrase Model}

We require sufficient diversity of paraphrase model so that it is able to introduce enough changes in the paraphrase of each round. The VAE \citep{kingma2013auto,rezende2014stochastic} is a deep generative model that allows learning rich, nonlinear representations for high dimensional inputs. It can improve the diversity by sampling from latent space. \citet{bowman-etal-2016-generating} proposed a new model to apply VAE to natural language generation for the first time. Our paraphrase model is based on conditional VAE with LSTM. Transformer \citep{vaswani2017attention} has achieved excellent performance in many tasks. But our experiments show that it may cause KL divergence to become 0, called posterior collapse, which means a decrease of diversity. So we do not employ Transformer as encoder and decoder.

We define the embedding matrices of $S_i$ and $P$ as $E_s^i = \{e_s^1, e_s^2, \cdots, e_s^{L_i}\}$ and $E_p = \{e_p^1, e_p^2, \cdots, e_p^{M}\}$ respectively, where $e_s^i,e_p^j \in \mathbb{R}^{d_e}$ are the embedding vector of the word in $S_i$ and $P$, and $d_e$ is the embedding dimension.

\subsubsection{Encoder}
Conditional VAE contains two encoders that share parameters: an original sentence encoder and a paraphrase encoder.
We first send $E_s^i$ into original sentence encoder to get its encoding matrix $O_s^i \in \mathbb{R}^{d_h \times L_i}$ and vector representation $h_s^i \in \mathbb{R}^{d_h}$ of $S_i$, where $d_h$ is the hidden dimension of LSTM. Then we send $E_p$ and $h_s^i$ into paraphrase encoder to get its vector representation $h_z$. $h_z$ is passed through two different feed-forward neural networks with parameter $\Phi$ to produce the mean $\mu$ and the variance $\sigma^2$ of the distribution of latent space. We can get the latent code $\bm{z} \in \mathbb{R}^{d_z}$ by sampling from latent space and reparameterization, where $d_z$ is the dimension of latent code.

\subsubsection{Decoder}
We define the embedding matrix which be sent into decoder as $E_d = \{e_d^1, e_d^2, \cdots, e_d^N\} \in \mathbb{R}^{d_e \times N}$. Then, we concatenate $z$ with the embedding vector $e_d^i$ as the input of decoder. The decoder also takes $h_s^i$ as input. The output of decoder is defined as $O_d^i \in \mathbb{R}^{d_h \times N}$. Then, an attention \citep{luong2015effective} and copy mechanism \citep{see2017get} are leveraged as follow.
First, we get the attention weight $p_a$ and attention vector $V_a$ as follow. 
\begin{equation}
\begin{aligned}
    p_a &= \operatorname{softmax}(O_d^i {O_s^i}^\top)\\
    &V_{a} = p_{a} O_s^i
\end{aligned}
\end{equation}

Then, we leverage them to calculate the decoder probability $p_d$ and copy probability $\eta$.
\begin{equation}
\begin{aligned}
    p_d &= \operatorname{softmax}(\mathbf{W}_{o}[O_d^i||V_{a}] + \mathbf{b}_o) \\
    \eta = \sigma(&\mathbf{W}_h [O_d^i||V_{a}] + \mathbf{W}_s \{e_d^i||\bm{z}\}_{i=1}^N + \mathbf{b}_{\eta}) \\
\end{aligned}
\end{equation}

\noindent where $\mathbf{W}_o, \mathbf{W}_h, \mathbf{W}_b, \mathbf{b}_o, \mathbf{b}_\eta$ are learnable parameters. $||$ is the concatenation operation. $\sigma$ is the sigmoid activation function. The final output probability of decoder is as follow.
\begin{equation}
    p = \eta p_d + (1 - \eta) p_a
\end{equation}

\subsubsection{Loss Function of Paraphrase Model}

The VAE with parameter $\Theta$ is trained by minimizing the following objective:

\begin{equation}
\begin{aligned}
    \mathcal{L}_{Para} = & -\operatorname{KL}(q_\Phi(\bm{z}|S_i, P)||p(\bm{z})) \\ &+ \mathbb{E}_{q_\Phi(\bm{z}|P)}[\operatorname{log}p_\Theta(P|\bm{z}, S_i)]
\label{elbo}
\end{aligned}
\end{equation}

\noindent where KL stands for the KL divergence. Eq. \ref{elbo} is called evidence lower bound, which provides a lower bound of $\log p(P|S_i;\Theta)$.

\citet{bowman-etal-2016-generating} figured out that variational inference for text generation often yields models that ignore their latent variables, a phenomenon called posterior collapse. This may cause the low diversity of generated sentences. To tackle this problem, we propose a diversity loss. We find that the diversity of the generated sentence is affected by its first word. For example, the first word can determine the form of a question sentence. Unfortunately, compared with the questions beginning with ``Is, May, Would'', we are more likely to collect questions beginning with ``What, When, How''. This can lead to serious category imbalances when generating the first word. So we set the penalty coefficient of the first-word loss as follow.

\begin{equation}
\begin{aligned}
\mathcal{L}_{w_1} &= \ln\left(\frac{N_b}{n_{w_1}}e\right) \log p(w_1|\Theta) \\
\end{aligned}
\end{equation}

\noindent where $N_b$ is the batch size during the training process, $n_{w_1}$ is the number of sentences beginning with $w_1$ in this batch. $e$ is the Euler's number that can make sure the penalty coefficient always no less than 1.

\subsection{Back-Translation Model}
Back-translation model aims to make sure the semantics of the generated paraphrase are the same with the original sentence during multi-round generation. It translates $S_i$ back to $S_0$. Different from paraphrase model which needs diversity, back-translation model is more focused on semantics maintaining. 
We employ Transformer \citep{vaswani2017attention} with copy mechanism as back-translation model because of its excellent performance in many tasks. 

The loss function of back-translation model is as follow: \footnote{Note that we also use $S_0$, $S_i$ and $P$  to denote the one-hot matrix of corresponding sentences.}:

\begin{equation}
\begin{aligned}
\mathcal{L}_s^i &= \operatorname{CrossEntropy}(\operatorname{BTModel}(S_i), S_0)\\
\mathcal{L}_p &= \operatorname{CrossEntropy}(\operatorname{BTModel}(P), S_0)\\
\mathcal{L}_{BT} &= \mathcal{L}_p + \lambda \sum_i \mathcal{L}_s^i
\end{aligned}
\label{loss_bt}
\end{equation}

\noindent where $\lambda$ is a hyper-parameter. $\operatorname{BTModel}$ indicates the back-translation model.

There are two parts in the loss of back-translation model: $\mathcal{L}_s^i$ and $\mathcal{L}_p$. We assume the $i$-th round paraphrase can be translated back to the original sentence $S_0$ if its semantics are preserved and thus we optimize $\mathcal{L}^i_s$. Similarly, the paraphrase $P$ can be translated back to the original sentence $S_0$ as well, so we also leverage $\mathcal{L}_p$ to train back-translation model. This can improve the generalization ability of the back-translation model, because back-translation model tends to guide paraphrase model to copy original sentence without changes if we do not employ true paraphrase data to train it.

\subsection{Gumble-Softmax Embedding}

We employ gumble-softmax embedding to connect each module of our model. We first define an embedding operation as follow:

\begin{equation}
    \operatorname{Embed} (X) = \mathbf{W}_e X
\end{equation}

For the probability $p$ generated by paraphrase model, we leverage gumble-softmax\cite{jang2017categorical} to get its one-hot matrix without sampling from multinomial distribution. Then we can get the embedding matrix $E$ as follow:

\begin{equation}
\begin{aligned}
    \operatorname{GS}(\bm{\pi}) &= \operatorname{softmax}((\operatorname{log}(\pi)_i + g_i) / \tau) \\
    E &= \operatorname{Embed}\left(\operatorname{GS}(p)\right)
\end{aligned}
\end{equation}

\noindent where $\bm{\pi}$ is a multinominal distribution wih $k$ dimension, $g_1, g_2, \cdots, g_i$ are i.i.d samples drawn from $\operatorname{Gumbel}(0, 1)$. $\tau$ is a hyper-parameter.

\begin{figure}[htbp]
\subfigure[] 
{
	\begin{minipage}{3.5cm}
	\centering          
	\includegraphics[scale=0.42]{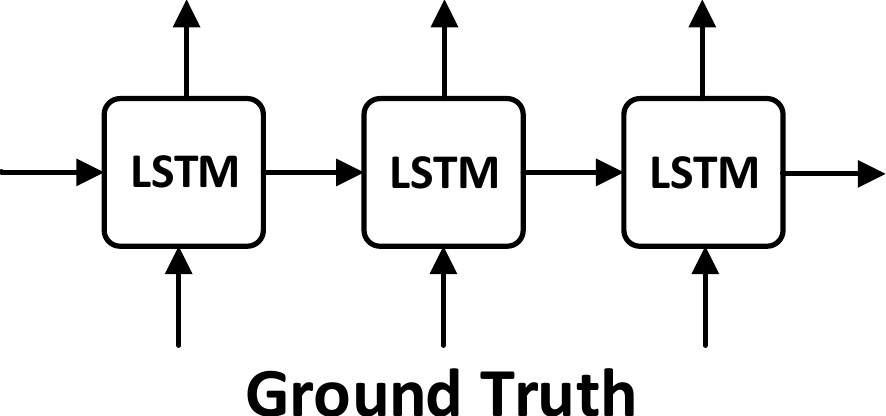}  
	\end{minipage}
}
\subfigure[]
{
	\begin{minipage}{3.5cm}
	\centering      
	\includegraphics[scale=0.42]{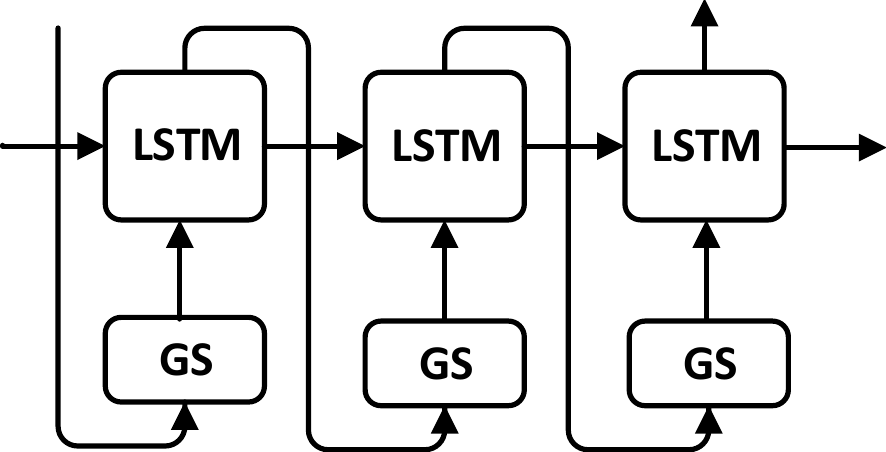}  
	\end{minipage}
}

\caption{Figure (a) shows the decoder with teacher forcing in the first round generation. Figure (b) shows the decoder with autoregression in other-round generation.} 
\label{gs} 
\end{figure}

There are three places in our model needing to leverage gumble-softmax embedding. First, we leverage it to embed the output probability of the paraphrase model as the input of the next-round paraphrase model. Next, gumble-softmax embedding is also used to connect the back-translation model with the paraphrase model. Figure \ref{overview} shows these two cases.  Finally, it is used in the multi-round paraphrase generation process to replace the teacher forcing. Generally, Seq2Seq model employs teacher forcing for model training, with using ground truth to guide the generation process. However, there is no ground truth in multi-round paraphrase generation, it can only generate sentence with a autoregressive method. We employ gumble-softmax to replace sampling in each step of the autoregressive process. Figure \ref{gs} shows this process.

\subsection{Loss Function}

We train paraphrase model together with back-translation model. The total loss of our model is as follow:

\begin{equation}
\begin{aligned}
    \mathcal{L} = \mathcal{L}_{para} + \mathcal{L}_{BT}
\end{aligned}
\end{equation}

Although we define a multi-round paraphrase model, we only train the first two rounds. Because we find that training too many rounds requires large computing resources, but can not improve the model performance significantly.
During inference, we can generate paraphrase more than two rounds.

\section{Experiment}

\subsection{Datasets}

We evaluate our BTmPG model on two benchmark datasets: 

\textbf{MSCOCO}\footnote{\url{https://cocodataset.org/}} \citep{lin2014microsoft} dataset contains human annotated captions of over 120k images. Each image contains five captions from five different annotators. This dataset has been widely used in previous works \citep{prakash2016neural,gupta2017deep,cao-wan-2020-divgan}.
We sample the MSCOCO according to \citet{prakash2016neural}.

\textbf{Quora}\footnote{\url{https://www.kaggle.com/c/quora-question-pairs/data?select=train.csv.zip}} dataset is a question paraphrase dataset. It contains over 400k question pairs. Each pair marked with a binary value indicates whether the questions in the pair are truly a duplicate of each other. So we select all such question pairs with binary value 1 as paraphrase dataset. There are about 150k question pairs in total. We randomly divide the training, validation and the test set. 

Table \ref{statistic} provides statistics of these two benchmark datasets.

\begin{table}[htb]
\centering
% \small
\scalebox{0.9}{
\begin{tabular}{c|ccc}
\toprule[1.2pt]

\textbf{Dataset} & \textbf{Train Set} &\textbf{Valid Set} & \textbf{Test Set}\cr
\hline
MSCOCO & 206,852 & 3,000 & 3,000\cr
Quora & 129,263 & 3,000 & 3,000\cr
\bottomrule[1.2pt]
\end{tabular}}
\caption{Statistic for datasets: the sizes of training, validation and test set.}
\label{statistic}
\end{table}

\subsection{Evaluation Metrics}

We use five widely-used metrics to evaluate paraphrases: BLEU4, self-BLEU, self-TER, BERTScore and p-BLEU.

\textbf{BLEU4} is widely used in generation tasks. It can measure how well the sentences generated by our model can match the references. Notice that some works also calculate the ROUGE\citep{lin-2004-rouge} or METEOR, but we think the role of these two metrics overlaps with BLEU4, as they all calculate the overlap degree between outputs and references. Therefore we only calculate BLEU4 to evaluate the match degree between outputs and references.

We evaluate the difference between the output sentence and the original sentence with two metrics. One of them is \textbf{self-BLEU} which is the BLEU4 score between the output sentence and the original sentence. The lower the value of self-BLEU, the more difference between output sentences and original sentences. Another is \textbf{self-TER}\footnote{We use the tool at \url{https://github.com/jhclark/multeval}.}. \textbf{TER}\citep{zaidan2010predicting} is used to evaluate the edit distance between two sentences. Self-TER is calculated as the TER between the output sentence and the original sentence.

\textbf{BERTScore} \footnote{The tool of BERTScore is available at \url{https://github.com/Tiiiger/bert_score}} is proposed by \citet{zhang2019bertscore} to evaluate the semantic similarity between the output sentence and the original sentence. 
BERTScore has been widely leveraged to measure semantic preserving in the paraphrase generation task \citep{cao-wan-2020-divgan}. However, there may be some problems for BERTScore on our task due to the low score for reference. This is because BERTScore is not perfect in measuring semantic relevance. But as far as we know, there is no better score to evaluate semantic preserving, so we report BERTScore as a reference for semantic preserving. More evaluation about semantic relevance is shown in human evaluation.

We leverage \textbf{p-BLEU} \citep{cao-wan-2020-divgan} to evaluate the difference between outputs in different rounds. Concretely, for outputs in $k$ rounds $\{y_1, y_2, \cdots, y_k\}$, the p-BLEU can be calculated as follow.

\begin{equation}
% \centering
\begin{aligned}
    \operatorname{p-BLEU} = \frac{\sum_i\sum_{j\neq i} \operatorname{BLEU4}(y_i, y_j)}{k \times (k-1)}
\end{aligned}
\end{equation}

The lower p-BLEU means higher diversity between outputs in different rounds.

Notice that, BLEU4 may not suitable for our task , because we focus on the diversity of paraphrase. BLEU4 can only measure the match degree between outputs and references. However, a sentence usually has many more reference paraphrases, while the target given in the dataset is only one reference. So we also perform human evaluation to evaluate the semantic relevance, readability and diversity of generated paraphrases. 

\subsection{Baseline}

As our model focuses on the diversity of paraphrase, we mainly compare our model with VAE-SVG-eq \citep{gupta2017deep}, DiPS\citep{dips2019}\footnote{The code is available at \url{https://github.com/malllabiisc/DiPS}.}, SOW-REAP\citep{goyal2020neural}\footnote{The code is available at \url{https://github.com/tagoyal/sow-reap-paraphrasing}.} and the decoding method proposed by \citet{thompson2020paraphrase}\footnote{DNPG \citep{li2019decomposable}, which controls semantics through encoding different levels of granularity respectively, can also enhance diversity. But the code and outputs are not provided, so we are not able to use it as baseline.}. The last method penalizes the n-gram appearing in the original sentence to make the paraphrase different from the original sentence and enhance diversity. We mark this method as N-gram Penalty. We employ two different hyper-parameters provided by the authors: one of them is low penalty for N-gram, and another is high penalty. In addition, we also compare our model with Transformer and Transformer copy.

\subsection{Training Details}

For both datasets, we truncate all the sentences longer than 20 words and maintain a vocabulary size of 25k. During testing, we replace UNK with the original word with the highest copy probability. 

For paraphrase model,  we leverage 2-layer LSTM. We set the embedding dimension $d_e$ to 300, hidden size $d_h$ of LSTM to 512. We set the latent code dimension $d_z$ to 128. For back-translation model, we leverage Transformer-copy with 3-layer encoder and decoder. We set the model size to 450, and the head number of multi-head attention to 9. We set $\lambda$ to 1, which will be discussed in our ablation study. For the hyper-parameter $\tau$ in gumble-softmax, we refer \citep{nie2018relgan} to increase the $\tau$ over iterations via an exponential policy: $\tau = \tau_{\max}^{-n_e/N_e}$, where $n_e$ is the current epoch and $N_e$ is the total number of epoch. We set $\tau_{\max}$ to 5. We train our model for 30 epochs. We set batch size to 50, and we select the model of the final epoch to generate paraphrase in test set.

\section{Result}
\subsection{Automatic Evaluation}
\begin{table*}[htb]
    \centering
    % \small
    \scalebox{0.73}{
    \begin{tabular}{lc|cccc|cccc}
    \toprule[1.2pt]
    \multicolumn{2}{c|}{\multirow{2}{*}{\textbf{Model}}} 
    & \multicolumn{4}{c|}{\textbf{MSCOCO}}   & \multicolumn{4}{c}{\textbf{Quora}} \cr
    \cline{3-10}
    && \textbf{BLEU4} & \textbf{self-BLEU} $\downarrow$ & \textbf{self-TER} $\uparrow$ &\textbf{BERTScore} $\uparrow$ & \textbf{BLEU4} & \textbf{self-BLEU} $\downarrow$ & \textbf{self-TER} $\uparrow$ &\textbf{BERTScore} $\uparrow$ \cr
    \hline
    \multicolumn{2}{l|}{Reference}          & -     & 8.12  & 78.40 & 46.20 & -     & 31.46 & 54.92 & 67.37\cr
    \multicolumn{2}{l|}{VAE-SVG-eq}         & 25.07 & 13.77 & 66.92 & 51.72 & 22.52 & 36.05 & 50.25 & 67.06\cr
    \multicolumn{2}{l|}{Transformer}        & 25.81 & 14.92 & 65.74 & 54.47 & 26.22 & 37.99 & 46.42 & 67.53\cr
    \multicolumn{2}{l|}{Transformer copy}   & 26.80 & 17.94 & 62.49 & 56.49 & 28.97 & 43.69 & 42.52 & 71.05\cr
    \multicolumn{2}{l|}{DiPS}               & 23.52 & 12.23 & 67.31 & 51.40 & 23.38 & 29.24 & 54.41 & 63.09\cr
    \multicolumn{2}{l|}{SOW-REAP}           & 15.31 & 44.22 & 39.42 & \textbf{63.68} & 15.36 & 47.62 & 38.98 & 62.21\cr
    \multicolumn{2}{l|}{N-gram Penalty-low} & 25.37 & 12.16 & 66.66 & 53.47 & 26.00 & 36.65 & 47.31 & 66.93\cr
    \multicolumn{2}{l|}{N-gram Penalty-high}& 23.68 & 0.00$^*$&69.24& 52.08 & 17.53 & 0.00$^*$&\textbf{59.30}& 59.20\cr

    \hline
    \multirow{3}{*}{\makecell[l]{BTmPG\\(Ours)}}
    &R1  &  25.54 & 18.50 & 61.78 & 59.34 & 28.02 & 58.47 & 33.99& \textbf{77.21} \cr
    % &R2&  25.57 & 15.23 & & 56.68 & 25.90 & 46.41 && 72.03 \cr
    % &R3&  24.71 & 13.90 & & 55.41 & 24.61 & 42.41 && 69.23 \cr
    % &R4&  24.08 & 13.19 & & 54.65 & 23.71 & 39.81 && 67.32 \cr
    &R5&  23.65 & 12.58 & 68.27 & 54.07 & 23.15 & 37.89 & 48.62 & 65.90 \cr
    &R10& 22.42 & \textbf{10.98} & \textbf{70.10} & 52.37 & 22.17 & \textbf{34.15} & 53.34& 62.91 \cr
    \bottomrule[1.2pt]
    \end{tabular}}
    
    \caption{\label{result} Automatic evaluation results on MSCOCO and Quora test sets. In the table, R1, R5 and R10 mean the first round, the fifth round and the tenth round of paraphrase generation.}
\end{table*}

Table \ref{result} shows the results of automatic evaluation. Our model substantially improves the BERTScore in the first round of paraphrase generation and generally gets the state-of-the-art performance. The value of self-BLEU can be significantly reduced with the increase of the round number of paraphrase generation while maintaining semantics.

For both datasets, the first round paraphrase generation of our model achieves the highest BERTScore than any other models. This is because back-translation model can provide sufficient semantic guidance for paraphrase model. As the increase of the round number, the values of self-BLEU and self-TER are reduced significantly, which means the paraphrase sentences our model generated are more and more different from original sentences. While BERTScore can still maintain a relatively high value. (A slight reduction of BERTScore is acceptable as BERTScore is not perfect in measuring semantic relevance.) We find that the paraphrase generated in the fifth round is good with balancing the diversity and the relevancy.

DiPS gets the BERTScore similar to round 5 generation, while its outputs lack of diversity compared with our model. SOW-REAP gets the highest BERTScore for MSCOCO, but it does not perform well on self-BLEU. Because SOW-REAP tends to generate paraphrase without change, the semantics of the paraphrase may be similar with the original sentence but the paraphrase lacks of diversity. N-gram Penalty with high penalty can lead self-BLEU to 0, as it strictly does not allow to generate those 4-grams appearing in the original sentence. Although the N-gram Penalty method can generate outputs totally different from original sentences, it fails to preserve the major semantics. However, our BTmPG model can increase diversity as much as possible while preserving major semantics.

To explore the pairwise diversity of our model's outputs in different rounds, we also calculate the p-BLEU values for VAE-SVG-eq and our model (p-BLEU is not suitable for other models). For VAE-SG-Eq, we generate 10 outputs by random sampling the latent space. For our model, we select the first 10 rounds outputs. Table \ref{p-bleu} shows the results of p-BLEU. The p-BLEU value of our model is much lower than VAE-SVG-eq, which means that our model has better ability to generate multiple diversified paraphrases than VAE-SVG-eq.

\subsection{Ablation Study}

\begin{figure}[htb]
\centering
\includegraphics[scale=0.47]{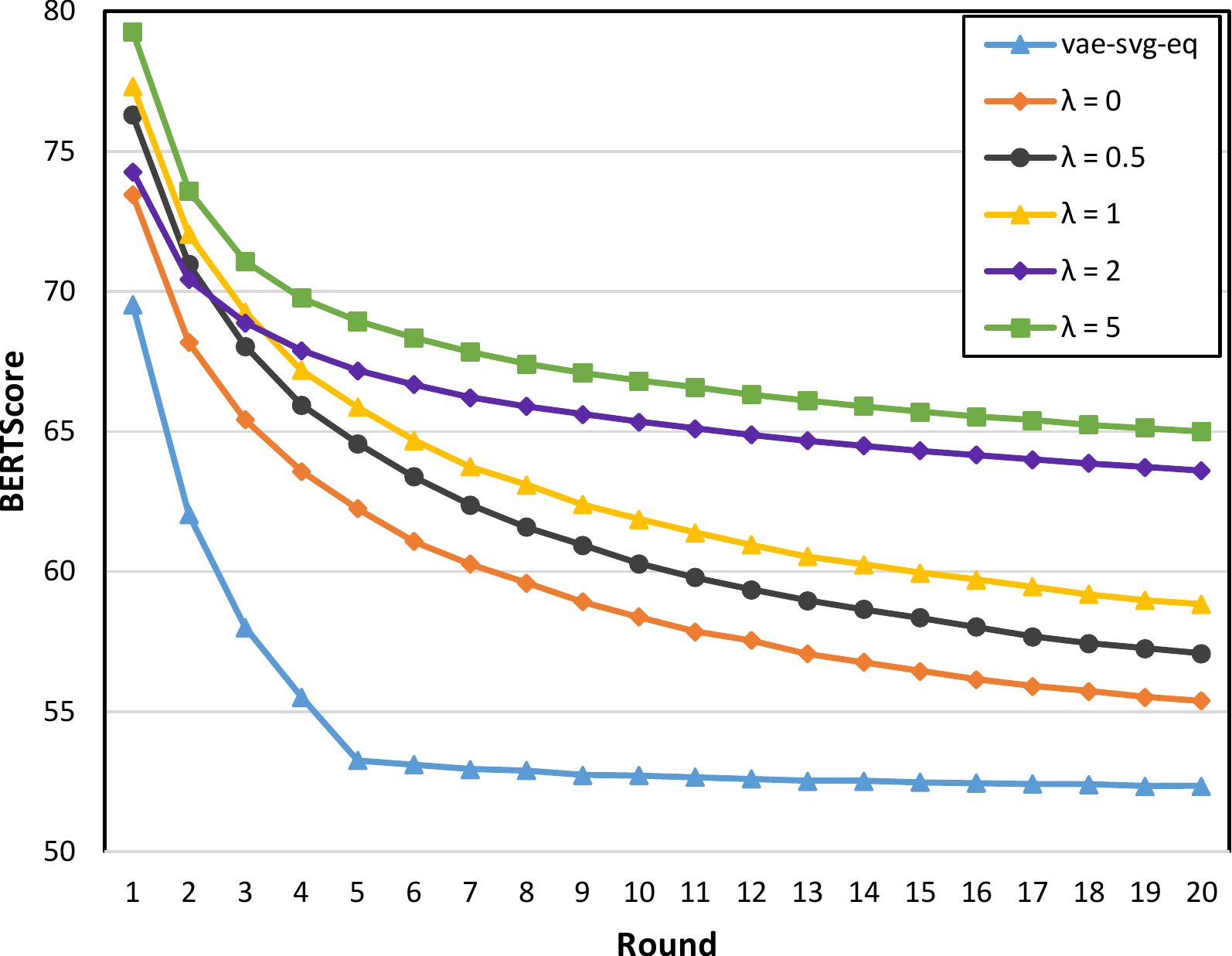}
\caption{ The BERTScore of paraphrases of 20 rounds on Quora.}
\label{abs}
\end{figure}

In this section, we will explore the role of back-translation model in preserving semantics. We set the hyper-parameter $\lambda$ from 0 to 5. A bigger $\lambda$ means back-translation provides more semantic guidance to paraphrase model. $\lambda = 0$ means that we remove back-translation model totally. We generate paraphrases of 20 rounds and calculate the values of BERTScore. In order to explore the effect of leveraging other paraphrase model in the multi-round generation framework, we also adopt VAE-SVG-eq in a multi-round generation process to generate paraphrases of 20 rounds on Quora, and compute the values of BERTScore. Figure \ref{abs} shows the trend of BERTScore with the increase of the round number.

\begin{table}
    \centering
    \small
    \scalebox{0.95}{
    \begin{tabular}{c|cc}
        \toprule[1.2pt]
        \multirow{2}{*}{\textbf{Model}} &  \multicolumn{2}{c}{\textbf{p-BLEU} $\downarrow$ }\cr
        \cline{2-3}
        &\textbf{MSCOCO}&\textbf{Quora} \cr
        \hline
        VAE-SVG-eq & 75.52 & 81.50 \cr
        BTmPG(Ours)& \textbf{62.83} & \textbf{67.60} \cr
        \bottomrule[1.2pt]
    \end{tabular}}
    \caption{The p-BLEU score for VAE-SVG-eq and BTmPG}
    \label{p-bleu}
\end{table}

Obviously, compared with VAE-SVG-eq, our improved VAE model can preserve semantics better. Back-translation can much improve the lower bound of BERTScore , which means back-translation can help to preserve the semantics during multi-round paraphrase generation. 

We also calculate the p-BLEU for the paraphrases of the first 10 rounds for different $\lambda$. Table \ref{lpbleu} shows the result. From the table we can know that, although back-translation can help to preserve semantics, a higher $\lambda$ can lead to a lack of diversity of paraphrase. Therefore, it is wise to select an appropriate $\lambda$ according to the actual requirement.

\begin{table}[htb]
    \centering\scalebox{0.9}{
    \begin{tabular}{c|ccccc}
       \toprule[1.2pt]
        $\bm{\lambda}$ & 0 & 0.5 & 1 & 2 & 5\cr
        \hline
        \textbf{p-BLEU} & 63.53 & 66.40 & 67.61 & 74.83 & 88.05 \cr
        \bottomrule[1.2pt]
    \end{tabular}}    
    \caption{The p-BLEU of paraphrases of the first 10 rounds for different $\lambda$.}

    \label{lpbleu}
\end{table}

\subsection{Human Evaluation}

We  perform  human  evaluation  on  system  outputs  with  respect to three aspects: relevancy, fluency and diversity. Relevancy indicates if the semantics of outputs and original are identical. Fluency indicates the readability of output sentences. Diversity indicates the lexical and syntactic differences between output sentences and original sentences and thus we use two indicators for lexical diversity and syntactic diversity respectively. 

We randomly sample 100 sentences from each test set and get a total of 200 sentences for evaluation. We employ 6 graduate students to rate each instance. We ensure every instance is rated by at least three judges. Table \ref{human} shows the result of human evaluation.

\begin{table}[htb]
    \centering
    % \small
    \scalebox{0.75}{
    \begin{tabular}{lc|cc|cc}
    \toprule[1.2pt]
    \multicolumn{2}{c|}{\multirow{2}{*}{\textbf{Model}}} & \multirow{2}{*}{\textbf{Relevancy}} &  \multirow{2}{*}{\textbf{Fluency}} &\multicolumn{2}{c}{\textbf{Diversity}} \cr
    % \cline{5-6}
    & & & & \textbf{Lexical} & \textbf{Syntactic} \cr
    \hline
    \multicolumn{2}{l|}{VAE-SVG-eq}                & 3.24  & 3.44 & 3.93 & 4.01 \cr
    \multicolumn{2}{l|}{Transformer}               & 3.82  & \textbf{3.96} & 3.71 & 3.77 \cr
    \multicolumn{2}{l|}{DiPS}                      & 3.62  & 3.50 & 3.64 & 3.70 \cr
    \multicolumn{2}{l|}{SOW-REAP}                  & 3.59  & 3.34 & 2.79 & 3.88 \cr
    \multicolumn{2}{l|}{N-gram Penalty}            & 3.44  & 3.65 & 3.79 & 3.68 \cr
    \hline
    \multirow{3}{*}{\makecell[l]{BTmPG\\(Ours)}}
    &R1                                            & \textbf{4.12}  & 3.92 & 3.65 & 3.85 \cr
    &R5                                            & 3.93  & 3.81 & 3.95 & 4.00 \cr
    &R10                                           & 3.84  & 3.82 & \textbf{4.20} & \textbf{4.15} \cr
    \bottomrule[1.2pt]
    \end{tabular}}
    \caption{Human evaluation results.}
    \label{human}
\end{table}

    From the table, we can see that the paraphrase in the first round can preserve more semantics of original sentence but lack of diversity. With the increase of the round number, the relevancy score decreases slightly, but the diversity scores increase substantially. Fluency may be influenced by diversity, because human may feel a slight decrease of fluency with the increase of diversity. As compared with other models, our model can generate paraphrases with high diversity, while maintaining semantics and fluency well. Previous models like SOW-REAP and DiPS can not maintain the semantics, though they can produce paraphrases with relatively high diversity.

\subsection{Case Study}

We perform case studies for better understanding the model performance. Table \ref{casestudy} shows an example of Quora, which include paraphrases of the first 15 rounds.

\begin{table}[htb]
    \centering
    \small
    \scalebox{0.93}{
    \begin{tabular}{|l|p{6cm}|}
        \hline
        \multicolumn{2}{|c|}{\textbf{Cases from Quora}}\\
        \hline
         Original & why did modi scrap rs 500 \& rs 1000 notes ? and what 's the reason for the sudden introduction of the 2000 rupee note ? \\
        \hline
         Reference & why did \textcolor{red}{goi demobilise} 500 and 1000 rupee notes ? \\
        \hline
         Round1 & why did the \textcolor{blue}{indian government ban} the 500 and 1000 rupee notes and why is it {\color{blue}bringing to} ? \\
        \hline
         Round2 & \textcolor{blue}{what do you think about} the ban on 500 and 1000 \textcolor{blue}{denomination} notes in \textcolor{blue}{india} ? \\
        \hline
         Round4 & \textcolor{blue}{how do you see} the pm modi ’s \textcolor{blue}{move} of banning 500 and 1000 rupee \textcolor{blue}{currency} notes ?\\
        \hline
         Round5 & \textcolor{blue}{what do you think of the decision by the indian government to demonetize} 500 and 1000 rupee notes ? \\
        \hline
         Round9 & is modi 's \textcolor{blue}{decision on demonetization} of 500 and 1000 notes by public modi ?\\
         \hline
     Round11& \textcolor{blue}{was the decision by the indian government to demonetize} 500 and 1000 notes \textcolor{blue}{right or wrong} ? \\
         \hline
         Round12& \textcolor{blue}{would banning} notes of {\color{blue}denominations} 500 and 1000 \textcolor{blue}{help to curb the black money in india} ? \\
         \hline
         Round13& \textcolor{blue}{what will be the effects of} banning 500 and 1000 rupees \textcolor{blue}{on indian economy} ? \\
         \hline
         Round14& \textcolor{blue}{what are the advantage of} banning 500 and 1000 rupees \textcolor{blue}{in Indian} ? \\
         \hline
         Round15& \textcolor{blue}{what are the pros and corns of} banning 500 and 1000 rupees \textcolor{blue}{by indian government} ? \\
         \hline
    \end{tabular}}
    \caption{An example of Quora and the generated paraphrases in multiple rounds. The word in color means that it does not appear in the original sentence.}
    \label{casestudy}
\end{table}

This case shows how does our model modify sentences during multi-round paraphrase generation process. With the increase of round number, the difference between the generated paraphrase and the original sentence becomes larger, while the paraphrase still preserves the major semantics of the original sentence.

\section{Conclusion}

In this paper, we focus on improving the diversity of generated paraphrase, i.e., making the generated paraphrase much more different from the original sentence. We propose a multi-round paraphrase generation method \textbf{BTmPG} with the guidance of back-translation. Both automatic and human evaluation results show that our method can generate diverse paraphrase while maintaining semantics. Ablation study proves back-translation is very helpful to preserve semantics. In the future, we will explore other methods such as GAN, to improve paraphrase diversity. We will also test our method on more languages other than English. 

\section*{Acknowledgments}

This work was supported by National Natural Science Foundation of China (61772036), Beijing
Academy of Artificial Intelligence (BAAI) and Key Laboratory of Science, Technology and Standard
in Press Industry (Key Laboratory of Intelligent Press Media Technology). We appreciate the anonymous reviewers for their helpful comments. Xiaojun Wan is the corresponding author.

\bibliographystyle{acl_natbib}
\bibliography{acl2021}

\begin{thebibliography}{30}
\expandafter\ifx\csname natexlab\endcsname\relax\def\natexlab#1{#1}\fi

\bibitem[{An and Liu(2019)}]{DBLP:journals/corr/abs-1909-13827}
Zhecheng An and Sicong Liu. 2019.
\newblock \href {http://arxiv.org/abs/1909.13827} {Towards diverse paraphrase
  generation using multi-class wasserstein {GAN}}.
\newblock \emph{CoRR}, abs/1909.13827.

\bibitem[{Bhagat and Hovy(2013)}]{rahul2013squibs}
Rahul Bhagat and Eduard Hovy. 2013.
\newblock \href {https://doi.org/10.1162/COLI_a_00166} {{S}quibs: What is a
  paraphrase?}
\newblock \emph{Computational Linguistics}, 39(3):463--472.

\bibitem[{Bowman et~al.(2016)Bowman, Vilnis, Vinyals, Dai, Jozefowicz, and
  Bengio}]{bowman-etal-2016-generating}
Samuel~R. Bowman, Luke Vilnis, Oriol Vinyals, Andrew Dai, Rafal Jozefowicz, and
  Samy Bengio. 2016.
\newblock \href {https://doi.org/10.18653/v1/K16-1002} {Generating sentences
  from a continuous space}.
\newblock In \emph{Proceedings of The 20th {SIGNLL} Conference on Computational
  Natural Language Learning}, pages 10--21, Berlin, Germany. Association for
  Computational Linguistics.

\bibitem[{Cao and Wan(2020)}]{cao-wan-2020-divgan}
Yue Cao and Xiaojun Wan. 2020.
\newblock \href {https://doi.org/10.18653/v1/2020.findings-emnlp.218}
  {{D}iv{GAN}: Towards diverse paraphrase generation via diversified generative
  adversarial network}.
\newblock In \emph{Findings of the Association for Computational Linguistics:
  EMNLP 2020}, pages 2411--2421, Online. Association for Computational
  Linguistics.

\bibitem[{Fu et~al.(2019)Fu, Feng, and Cunningham}]{fu2019paraphrase}
Yao Fu, Yansong Feng, and John~P Cunningham. 2019.
\newblock Paraphrase generation with latent bag of words.
\newblock In \emph{Advances in Neural Information Processing Systems}, pages
  13645--13656.

\bibitem[{Goyal and Durrett(2020)}]{goyal2020neural}
Tanya Goyal and Greg Durrett. 2020.
\newblock \href {https://doi.org/10.18653/v1/2020.acl-main.22} {Neural
  syntactic preordering for controlled paraphrase generation}.
\newblock In \emph{Proceedings of the 58th Annual Meeting of the Association
  for Computational Linguistics}, pages 238--252, Online. Association for
  Computational Linguistics.

\bibitem[{Gupta et~al.(2018)Gupta, Agarwal, Singh, and Rai}]{gupta2017deep}
Ankush Gupta, Arvind Agarwal, Prawaan Singh, and Piyush Rai. 2018.
\newblock A deep generative framework for paraphrase generation.
\newblock \emph{Proceedings of the AAAI Conference on Artificial Intelligence}.

\bibitem[{Jang et~al.(2017)Jang, Gu, and Poole}]{jang2017categorical}
Eric Jang, Shixiang Gu, and Ben Poole. 2017.
\newblock Categorical reparameterization with gumbel-softmax.
\newblock \emph{International conference on learning representations}.

\bibitem[{Kingma and Welling(2013)}]{kingma2013auto}
Diederik~P Kingma and Max Welling. 2013.
\newblock Auto-encoding variational bayes.
\newblock \emph{International Conference on Learning Representations}.

\bibitem[{Kumar et~al.(2019)Kumar, Bhattamishra, Bhandari, and
  Talukdar}]{dips2019}
Ashutosh Kumar, Satwik Bhattamishra, Manik Bhandari, and Partha Talukdar. 2019.
\newblock \href {https://www.aclweb.org/anthology/N19-1363} {Submodular
  optimization-based diverse paraphrasing and its effectiveness in data
  augmentation}.
\newblock In \emph{Proceedings of the 2019 Conference of the North {A}merican
  Chapter of the Association for Computational Linguistics: Human Language
  Technologies, Volume 1 (Long and Short Papers)}, pages 3609--3619,
  Minneapolis, Minnesota. Association for Computational Linguistics.

\bibitem[{Li et~al.(2020)Li, Sha, and Shi}]{li2020revisiting}
Hongzheng Li, Jiu Sha, and Can Shi. 2020.
\newblock Revisiting back-translation for low-resource machine translation
  between chinese and vietnamese.
\newblock \emph{IEEE Access}, 8:119931--119939.

\bibitem[{Li and Specia(2019)}]{li2019improving}
Zhenhao Li and Lucia Specia. 2019.
\newblock \href {https://doi.org/10.18653/v1/D19-5543} {Improving neural
  machine translation robustness via data augmentation: Beyond
  back-translation}.
\newblock In \emph{Proceedings of the 5th Workshop on Noisy User-generated Text
  (W-NUT 2019)}, pages 328--336, Hong Kong, China. Association for
  Computational Linguistics.

\bibitem[{Li et~al.(2019)Li, Jiang, Shang, and Liu}]{li2019decomposable}
Zichao Li, Xin Jiang, Lifeng Shang, and Qun Liu. 2019.
\newblock \href {https://doi.org/10.18653/v1/P19-1332} {Decomposable neural
  paraphrase generation}.
\newblock In \emph{Proceedings of the 57th Annual Meeting of the Association
  for Computational Linguistics}, pages 3403--3414, Florence, Italy.
  Association for Computational Linguistics.

\bibitem[{Lin(2004)}]{lin-2004-rouge}
Chin-Yew Lin. 2004.
\newblock \href {https://www.aclweb.org/anthology/W04-1013} {{ROUGE}: A package
  for automatic evaluation of summaries}.
\newblock In \emph{Text Summarization Branches Out}, pages 74--81, Barcelona,
  Spain. Association for Computational Linguistics.

\bibitem[{Lin et~al.(2014)Lin, Maire, Belongie, Hays, Perona, Ramanan,
  Doll{\'a}r, and Zitnick}]{lin2014microsoft}
Tsung-Yi Lin, Michael Maire, Serge Belongie, James Hays, Pietro Perona, Deva
  Ramanan, Piotr Doll{\'a}r, and C~Lawrence Zitnick. 2014.
\newblock Microsoft coco: Common objects in context.
\newblock In \emph{European conference on computer vision}, pages 740--755.
  Springer.

\bibitem[{Liu et~al.(2020)Liu, Mou, Meng, Zhou, Zhou, and
  Song}]{liu-etal-2020-unsupervised}
Xianggen Liu, Lili Mou, Fandong Meng, Hao Zhou, Jie Zhou, and Sen Song. 2020.
\newblock \href {https://doi.org/10.18653/v1/2020.acl-main.28} {Unsupervised
  paraphrasing by simulated annealing}.
\newblock In \emph{Proceedings of the 58th Annual Meeting of the Association
  for Computational Linguistics}, pages 302--312, Online. Association for
  Computational Linguistics.

\bibitem[{Luong et~al.(2015)Luong, Pham, and Manning}]{luong2015effective}
Thang Luong, Hieu Pham, and Christopher~D. Manning. 2015.
\newblock \href {https://doi.org/10.18653/v1/D15-1166} {Effective approaches to
  attention-based neural machine translation}.
\newblock In \emph{Proceedings of the 2015 Conference on Empirical Methods in
  Natural Language Processing}, pages 1412--1421, Lisbon, Portugal. Association
  for Computational Linguistics.

\bibitem[{Nie et~al.(2019)Nie, Narodytska, and Patel}]{nie2018relgan}
Weili Nie, Nina Narodytska, and Ankit Patel. 2019.
\newblock \href {https://openreview.net/forum?id=rJedV3R5tm} {Relgan:
  Relational generative adversarial networks for text generation}.
\newblock In \emph{7th International Conference on Learning Representations,
  {ICLR} 2019, New Orleans, LA, USA, May 6-9, 2019}. OpenReview.net.

\bibitem[{Prakash et~al.(2016)Prakash, Hasan, Lee, Datla, Qadir, Liu, and
  Farri}]{prakash2016neural}
Aaditya Prakash, Sadid~A. Hasan, Kathy Lee, Vivek Datla, Ashequl Qadir, Joey
  Liu, and Oladimeji Farri. 2016.
\newblock \href {https://www.aclweb.org/anthology/C16-1275} {Neural paraphrase
  generation with stacked residual {LSTM} networks}.
\newblock In \emph{Proceedings of {COLING} 2016, the 26th International
  Conference on Computational Linguistics: Technical Papers}, pages 2923--2934,
  Osaka, Japan. The COLING 2016 Organizing Committee.

\bibitem[{Rezende et~al.(2014)Rezende, Mohamed, and
  Wierstra}]{rezende2014stochastic}
Danilo~Jimenez Rezende, Shakir Mohamed, and Daan Wierstra. 2014.
\newblock Stochastic backpropagation and approximate inference in deep
  generative models.
\newblock \emph{international conference on machine learning.}, pages
  1278--1286.

\bibitem[{See et~al.(2017)See, Liu, and Manning}]{see2017get}
Abigail See, Peter~J. Liu, and Christopher~D. Manning. 2017.
\newblock \href {https://doi.org/10.18653/v1/P17-1099} {Get to the point:
  Summarization with pointer-generator networks}.
\newblock In \emph{Proceedings of the 55th Annual Meeting of the Association
  for Computational Linguistics (Volume 1: Long Papers)}, pages 1073--1083,
  Vancouver, Canada. Association for Computational Linguistics.

\bibitem[{Siddique et~al.(2020)Siddique, Oymak, and
  Hristidis}]{10.1145/3394486.3403231}
A.~B. Siddique, Samet Oymak, and Vagelis Hristidis. 2020.
\newblock Unsupervised paraphrasing via deep reinforcement learning.
\newblock In \emph{Association for Computing Machinery}, KDD '20, page
  1800–1809.

\bibitem[{Thompson and Post(2020)}]{thompson2020paraphrase}
Brian Thompson and Matt Post. 2020.
\newblock \href {https://www.aclweb.org/anthology/2020.wmt-1.67} {Paraphrase
  generation as zero-shot multilingual translation: Disentangling semantic
  similarity from lexical and syntactic diversity}.
\newblock In \emph{Proceedings of the Fifth Conference on Machine Translation},
  pages 561--570, Online. Association for Computational Linguistics.

\bibitem[{Vaswani et~al.(2017)Vaswani, Shazeer, Parmar, Uszkoreit, Jones,
  Gomez, Kaiser, and Polosukhin}]{vaswani2017attention}
Ashish Vaswani, Noam Shazeer, Niki Parmar, Jakob Uszkoreit, Llion Jones,
  Aidan~N Gomez, {\L}ukasz Kaiser, and Illia Polosukhin. 2017.
\newblock Attention is all you need.
\newblock In \emph{Advances in neural information processing systems}, pages
  5998--6008.

\bibitem[{Wang et~al.(2019)Wang, Gupta, Chang, and Baldridge}]{wang2019task}
Su~Wang, Rahul Gupta, Nancy Chang, and Jason Baldridge. 2019.
\newblock A task in a suit and a tie: paraphrase generation with semantic
  augmentation.
\newblock In \emph{Proceedings of the AAAI Conference on Artificial
  Intelligence}, volume~33, pages 7176--7183.

\bibitem[{Wubben et~al.(2010)Wubben, van~den Bosch, and
  Krahmer}]{wubben2010paraphrase}
Sander Wubben, Antal van~den Bosch, and Emiel Krahmer. 2010.
\newblock \href {https://www.aclweb.org/anthology/W10-4223} {Paraphrase
  generation as monolingual translation: Data and evaluation}.
\newblock In \emph{Proceedings of the 6th International Natural Language
  Generation Conference}.

\bibitem[{Zaidan and Callison-Burch(2010)}]{zaidan2010predicting}
Omar~F. Zaidan and Chris Callison-Burch. 2010.
\newblock \href {https://www.aclweb.org/anthology/N10-1057} {Predicting
  human-targeted translation edit rate via untrained human annotators}.
\newblock In \emph{Human Language Technologies: The 2010 Annual Conference of
  the North {A}merican Chapter of the Association for Computational
  Linguistics}, pages 369--372, Los Angeles, California. Association for
  Computational Linguistics.

\bibitem[{Zhang et~al.(2020)Zhang, Kishore, Wu, Weinberger, and
  Artzi}]{zhang2019bertscore}
Tianyi Zhang, Varsha Kishore, Felix Wu, Kilian~Q. Weinberger, and Yoav Artzi.
  2020.
\newblock \href {https://openreview.net/forum?id=SkeHuCVFDr} {Bertscore:
  Evaluating text generation with {BERT}}.
\newblock In \emph{8th International Conference on Learning Representations,
  {ICLR} 2020, Addis Ababa, Ethiopia, April 26-30, 2020}. OpenReview.net.

\bibitem[{Zhao et~al.(2010)Zhao, Wang, Lan, and Liu}]{zhao2010leveraging}
Shiqi Zhao, Haifeng Wang, Xiang Lan, and Ting Liu. 2010.
\newblock \href {https://www.aclweb.org/anthology/C10-1149} {Leveraging
  multiple {MT} engines for paraphrase generation}.
\newblock In \emph{Proceedings of the 23rd International Conference on
  Computational Linguistics (Coling 2010)}, pages 1326--1334, Beijing, China.
  Coling 2010 Organizing Committee.

\bibitem[{Zhao et~al.(2020)Zhao, Chen, Chen, and Yu}]{zhao2020semi}
Yanbin Zhao, Lu~Chen, Zhi Chen, and Kai Yu. 2020.
\newblock Semi-supervised text simplification with back-translation and
  asymmetric denoising autoencoders.
\newblock In \emph{AAAI}, pages 9668--9675.

\end{thebibliography}

%\appendix

\end{document}